\documentclass{article}
\PassOptionsToPackage{authoryear}{natbib}
\usepackage[preprint]{neurips_2026}
\usepackage[utf8]{inputenc}
\usepackage[T1]{fontenc}
\usepackage{amsmath,amsfonts,amssymb}
\usepackage{graphicx}
\usepackage{booktabs,makecell,multirow,array,colortbl}
\usepackage{tabularx}
\usepackage{algorithm}
\usepackage{algpseudocode}
\usepackage{microtype}
\usepackage{xcolor}
\usepackage{xurl}
\usepackage{hyperref}
\usepackage{adjustbox}
\hypersetup{hidelinks}
\graphicspath{{figures/}{./}}
\newenvironment{keywords}{\par\noindent\textbf{Keywords:}\ }{\par}
\newcommand{\sep}{;\ }

\title{Rethinking the Tradeoff Between Temporal Encoding and Nonlinear Computation in Spiking Language Models}

\author{%
  \textbf{Hanfei Liu}$^{1,2}$ \quad
  \textbf{Shuchang Feng}$^{1,2}$ \quad
  \textbf{Yanxia Chen}$^{1,2}$ \\
  \textbf{Changzeng Fu}$^{1,2}$ \quad
  \textbf{Shiqi Zhao}$^{1,2}$\thanks{Corresponding author.} \\[3pt]
  $^{1}$Northeastern University, 110819, Shenyang, China \\
  $^{2}$Hebei Key Laboratory of Marine Perception Network \\
  and Data Processing, 066004, Qinhuangdao, China \\[3pt]
  \texttt{liuhf@mails.neu.edu.cn} \quad
  \texttt{fengshuchang@mails.neu.edu.cn} \\
  \texttt{chenyanxia@mails.neu.edu.cn} \quad
  \texttt{fuchangzeng@qhd.neu.edu.cn} \\
  \texttt{zhaoshiqi@qhd.neu.edu.cn}
}

\begin{document}
\maketitle

\begin{abstract}
Spiking language models face a tradeoff between representing continuous semantic features over short temporal windows and retaining costly nonlinear attention operations. We introduce Spora, which jointly designs spike encodings and attention operators. Binary temporal weights let $T$ spikes represent compositional values with up to $T$ bits of capacity, compared with $O(\log_2 T)$ bits for spike-count readout. Unipolar Binary Spiking (UBS) uses thresholds and spike-triggered residual decay to produce non-negative integer codes; Bipolar Binary Spiking (BBS) separates sign and magnitude and learns a scale for signed activations. These representations support accumulation-and-shift dot products and integer-exponent mappings in attention. With four time steps, Spora achieves 76.6 average GLUE score and 44.1 CoLA MCC, improving over SpikeLM by 1.2 and 6.2 points, respectively. Extending BBS to six steps raises these scores to 78.2 and 47.4. Conditional-decay analysis, matched-budget activation-quantization comparisons, event-workload statistics, and fixed-point evaluation further characterize the connection between encoding fidelity and computational cost.
\end{abstract}


\section{Introduction}

\begin{figure*}[t]
  \centering
  \includegraphics[width=\textwidth]{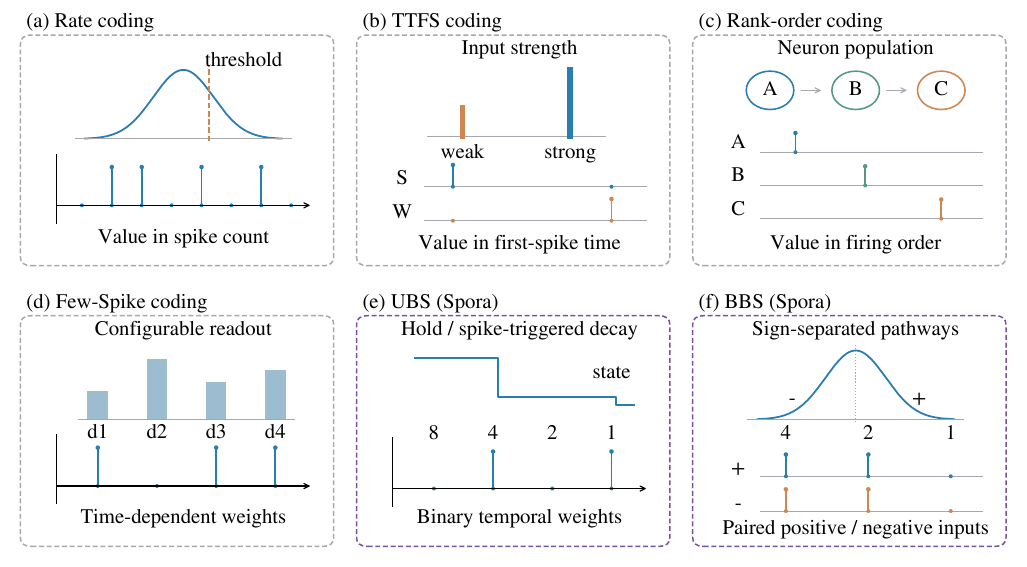}
  \caption{Spike encodings: (a) rate, (b) TTFS,
  (c) rank-order, (d) Few-Spike, (e) UBS, and (f) BBS.
  Event labels denote readout weights; BBS traces show
  separate positive and negative inputs.}
  \label{fig:neuron-mechanism}
\end{figure*}

Spiking Neural Networks (SNNs) process information through discrete events and membrane-potential dynamics inspired by biological neurons~\citep{mehonic2022brain,zhang2020system,maass1997networks}. Their sparse communication and event-driven computation offer a route to energy-efficient neural processing~\citep{NEURIPS2021_c4ca4238}. Recent advances in spiking vision~\citep{9543525,9597475}, particularly Transformer-based architectures~\citep{vaswani2023attentionneed,zhou2025spikingformerkeyfoundationmodel,zhou2023spikformer,LI2024127279,yao2023spikedriven}, have achieved competitive performance with four or fewer time steps.

Motivated by the generalization capabilities of large-scale language models~\citep{pmlr-v162-du22c}, SpikeLM~\citep{xing2024spikelmgeneralspikedrivenlanguage} and SpikeGPT~\citep{zhu2024spikegptgenerativepretrainedlanguage} demonstrate the feasibility of spike-driven language processing. Language representations, however, are dense, signed, and context-dependent. Transformer predictions depend on both the precision of these features and their interactions through attention. Mapping continuous activations into short spike sequences can therefore affect not only individual feature values but also the attention distributions that combine information across tokens.

Existing spiking language models approach this problem from two directions. Models that introduce spikes into selected linear modules retain floating-point Softmax to preserve attention fidelity~\citep{xing2024spikelmgeneralspikedrivenlanguage}. Hardware-oriented alternatives replace nonlinear operations with approximations~\citep{tang2025sorbet,WE-SpikingFormer}, while temporal budgets remain important for representation precision. Increasing the number of steps provides more encoding opportunities but also increases temporal processing. \textit{Although these two categories differ in implementation,
they face the same fundamental challenge: preserving fine-grained
semantic information within a short temporal window while avoiding costly nonlinear
attention computation.}

Figure~\ref{fig:neuron-mechanism} illustrates the relevant encoding choices. Rate coding counts events, TTFS uses first-spike latency~\citep{TTFS32}, and rank-order coding uses relative firing order across neurons~\citep{RankCode}. Weighted-spike and Few-Spike methods assign numerical readout weights to temporal events~\citep{KIM2018373,stockl2021optimized}. Our approach uses binary place values: an event's time index identifies its numerical contribution. Unlike a spike-count estimate, this readout forms an integer operand that can enter subsequent arithmetic directly.

Following this principle, Spora combines Unipolar Binary Spiking (UBS) and Bipolar Binary Spiking (BBS) encodings. UBS configures input partitions through thresholds and spike-triggered residual decay while retaining fixed binary readout weights. BBS separates sign from magnitude and uses a learnable scale to represent signed activations. These complementary codes support accumulation-and-shift dot products and integer-exponent attention mappings. {The design follows two requirements: encoding should adapt to activation distributions, and the resulting values should preserve the arithmetic structure needed by the next operator. Conditional decay and learned scales serve the first requirement; binary readout serves the second. This connects the temporal representation to the attention computation that consumes it.}

Our contributions are:

\textbf{(1) Unipolar encoding.}
{UBS distinguishes the numerical values specified by binary readout from the input intervals determined by thresholds and conditional decay. This provides short-window integer representations for shift-based exponential approximation.}

\textbf{(2) Bipolar encoding.}
BBS applies a shared magnitude process to both signs, with a learnable scale that adapts its finite signed codebook to the feature range.

\textbf{(3) Spiking attention and evaluation.}
The combined design replaces major attention multiplications and exponential evaluations with accumulation and shifts, yielding a theoretical 98.53\% reduction in {core attention energy}. Four-step Spora reaches 76.6 average GLUE score and 44.1 CoLA MCC, exceeding SpikeLM by 1.2 and 6.2 points. {Conditional-decay analysis, matched-budget Static-QAT comparisons, event workloads, and fixed-point experiments connect these task results to encoding fidelity and computational cost.}

\section{Related Work}

\textbf{Spiking neural encoding.}
Rate coding uses equally weighted events to approximate continuous values. Spike Firing Approximation~\citep{SFA} and memory-efficient training methods~\citep{NEURIPS2022_82846e19,Towards_Memory} reduce training costs, but spike-count readout still limits the number of representable levels. Temporal alternatives include TTFS~\citep{TTFS32,zhao2025ttfsformer} and rank-order coding~\citep{RankCode}, with tradeoffs in training stability and decoding dynamics~\citep{10378556}. Weighted-spike and Few-Spike methods introduce temporal numerical weights~\citep{KIM2018373,stockl2021optimized}. Spora uses fixed binary weights and conditional residual decay to obtain integer codes suited to signed features and attention operators.

\textbf{Transformer-based SNNs.}
Spikformer~\citep{zhou2023spikformer} replaces Softmax with spiking self-attention and reaches 74.81\% ImageNet-1k accuracy~\citep{ImageNet} in four steps. Spike-driven Transformer~\citep{yao2023spikedriven} uses additions and masking, while QKformer~\citep{QKFormer} introduces linear-complexity Q-K attention. MaxFormer~\citep{fang2026spiking} and SpikePool~\citep{lee2025spikepooleventdrivenspikingtransformer} use max-pooling in patch embedding and attention, respectively, to improve signal transmission. Language models such as BERT~\citep{devlin-etal-2019-bert} additionally require precise interactions between context-dependent token features, motivating joint treatment of representation and attention approximation.

\textbf{Hardware-adapted language SNNs.}
SpikeBERT~\citep{lv2024spikebert} uses Not-XOR spiking attention, and WE-SpikingFormer~\citep{WE-SpikingFormer} uses winner-take-all attention. Sorbet~\citep{tang2025sorbet} combines shift-based PTsoftmax and PowerNorm to avoid multiplication, division, and square roots, using sixteen time steps. SpikeLM~\citep{xing2024spikelmgeneralspikedrivenlanguage} improves information flow through ternary spikes and adaptive thresholds, but retains exponential attention operations, as does SpikeLLM~\citep{xing2025spikellm}. Spora instead links binary temporal readout to integer-exponent attention, addressing short-window precision and nonlinear computation within one encoding design.

\section{Method}

\begin{figure*}[tbp]
  \centering
  \includegraphics[width=0.90\linewidth]{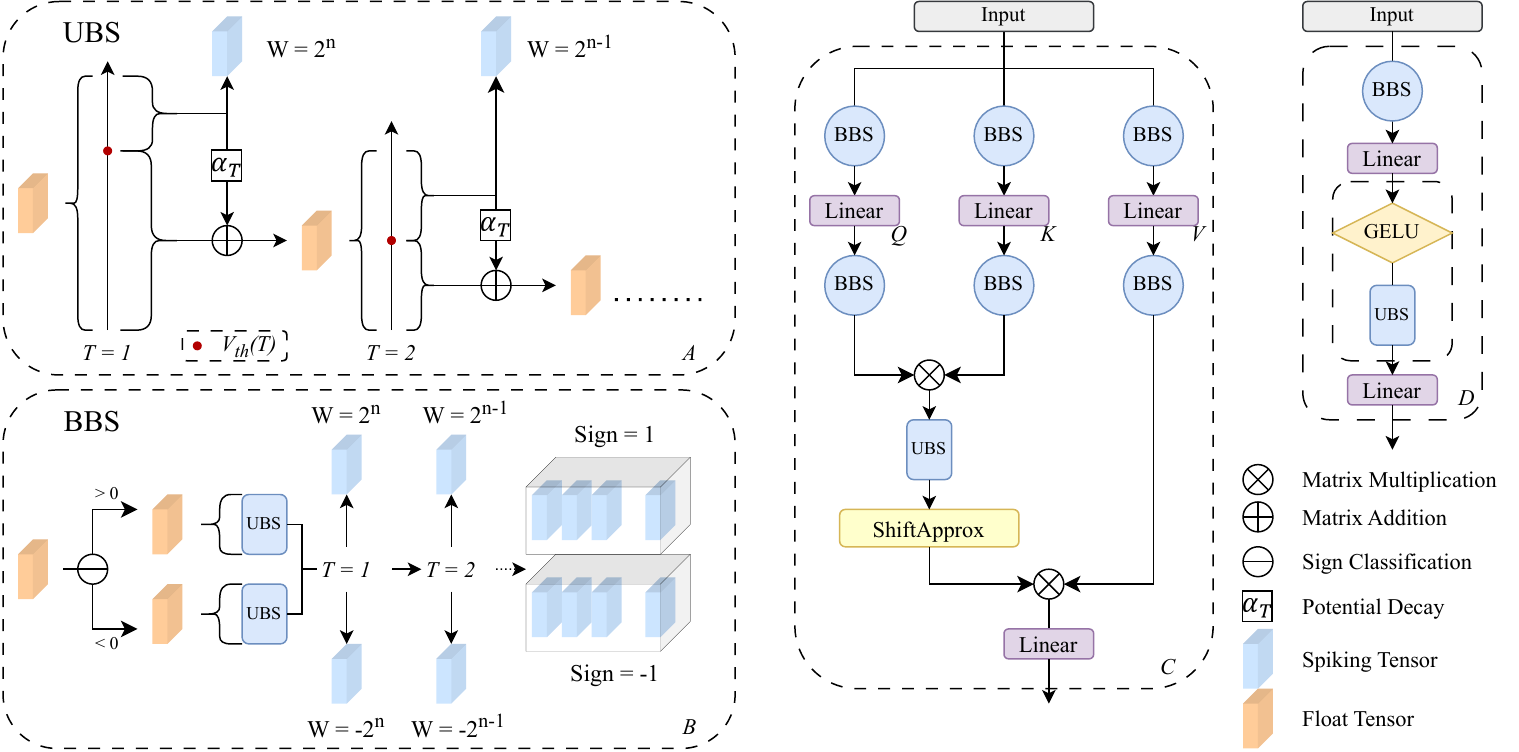}
  \caption{Spora architecture. (A) UBS temporal weights;
  (B) sign-separated BBS; (C) approximate attention, with
  ShiftApprox denoting shift-based Softmax approximation;
  (D) feed-forward block, where UBS implements the GELU
  approximation during inference.}
  \label{fig:method-overview}
\end{figure*}

Spora combines BBS-encoded signed features with UBS-based nonlinear mappings (Figure~\ref{fig:method-overview}). Binary temporal weights give both representations explicit numerical values for accumulation-and-shift computation.

\subsection{Representational Capacity and Numerical Roles}
\label{sec:capacity}

We measure the size of a decoded alphabet by $C=\log_2|\mathcal C|$, where $\mathcal C$ is the set of distinct readout values. For a binary sequence of length $U$, count-based decoding retains only $K=\sum_{t=1}^{U}S(t)$, whereas binary-weight decoding retains its place values:
\begin{equation}
\begin{aligned}
\mathcal C_{\mathrm{count}}&=\{0,\ldots,U\},\\
C_{\mathrm{count}}&=\log_2(U+1),\\
\mathcal C_{\mathrm{binary}}&=\{0,\ldots,2^U-1\},\\
C_{\mathrm{binary}}&=U.
\end{aligned}
\label{eq:capacity}
\end{equation}
For example, at $U=4$, count readout maps the sixteen possible spike patterns to five values, while binary readout assigns each pattern a distinct integer. This is an alphabet-capacity comparison: the thresholds and residual dynamics determine which patterns an input distribution actually occupies. UBS uses the larger alphabet for nonlinear mappings whose outputs must have explicit integer meaning.

BBS extends this construction to signed features. With $M$ magnitude decisions, the signed alphabet contains at most $2(2^M-1)+1=2^{M+1}-1$ states. Throughout the experiments, $T=2M$ denotes the BBS configuration with $M$ positive and $M$ negative bit planes. Thus, $T=4$ and $T=6$ correspond to seven and fifteen signed states. UBS uses $U=4$ stages in both configurations. This separates the BBS temporal budget from the UBS exponent-code budget.

\subsection{Unipolar Binary Spiking Encoding}

UBS processes larger-weight decisions before lower-weight refinements. We use $U$ for its number of stages, reserving $T$ for the BBS configuration reported in the experimental tables.

For non-negative input $x$, UBS initializes $V(0)=x$ and uses $U$ temporal weights $w(t)=2^{U-t}$. At each step, a threshold comparison generates a binary spike:
\begin{equation}
  \begin{aligned}
    S(t)&=\Theta\!\left[V(t-1)-V_{\mathrm{th}}(t)\right],\\
    S(t)&\in\{0,1\},
  \end{aligned}
  \label{eq:ubs-spike}
\end{equation}
where $\Theta$ is the Heaviside function. When $S(t)=1$, the threshold is subtracted and the remaining potential is multiplied by $\alpha$. When $S(t)=0$, the potential is passed unchanged to the next comparison. This spike-triggered update is
\begin{equation}
  V(t)=
  \begin{cases}
    \alpha\left[V(t-1)-V_{\mathrm{th}}(t)\right], & S(t)=1,\\
    V(t-1), & S(t)=0.
  \end{cases}
  \label{eq:ubs-update}
\end{equation}
The decoded value is $\mathrm{UBS}(x)=\sum_{t=1}^{U}w(t)S(t)$.
{\textbf{How conditional decay shapes the code.}
Binary readout specifies output values, while the residual dynamics determine their input regions. Consider the first two decisions, writing $\theta_t=V_{\mathrm{th}}(t)$. The second spike satisfies
\begin{equation}
  \begin{gathered}
    S(2)=1\quad\Longleftrightarrow\\
    \begin{cases}
      x\geq\theta_2, & S(1)=0,\\
      x\geq\theta_1+\theta_2/\alpha, & S(1)=1.
    \end{cases}
  \end{gathered}
  \label{eq:conditional-boundary}
\end{equation}
Thus, the effective threshold for a later event depends on whether an earlier event fired. Changing $\alpha$ moves this boundary without changing either event's readout weight. Over multiple steps, this mechanism shapes input partitions through the preceding spike sequence. The no-decay comparison in Section~\ref{sec:encoding-attention} tests the effect of this partitioning on attention fidelity while retaining binary readout.}

Thresholds and decay are fitted to the target mapping while keeping the readout weights fixed; Section~\ref{sec:training-encoding} gives the search procedure and training gradients.

\subsection{Bipolar Binary Spiking Encoding}

BBS separates sign and magnitude to encode signed activations, such as layer-normalized features, building on bidirectional spike representations in SpikeLM~\citep{xing2024spikelmgeneralspikedrivenlanguage}. For $x\in\mathbb{R}$, it initializes $V(0)=|x|/s$ with learnable scale $s>0$ and applies Equations~\ref{eq:ubs-spike}--\ref{eq:ubs-update} to the magnitude. With $M=T/2$ magnitude steps and binary weights $w_M(t)=2^{M-t}$, the output is
\begin{equation}
  \begin{aligned}
    Q(|x|)&=\sum_{t=1}^{M}w_M(t)S(t),\\
    \mathrm{BBS}(x)&=s\,\mathrm{sign}(x)\,Q(|x|).
  \end{aligned}
  \label{eq:bbs-quant}
\end{equation}
The $T$ binary spike tensors comprise $M$ positive and $M$ negative pathways; negative-path events contribute with a minus sign. The default $T=4$ configuration uses magnitude thresholds and weights $[2,1]$, while $T=6$ uses $[4,2,1]$. The scale adapts the fixed-threshold range to activation distributions and can be fused into the corresponding linear weights for accumulation-based inference.

Separating polarity from magnitude gives positive and negative inputs the same magnitude-encoding rule. At $T=4$, the decoded magnitudes range from 0 to 3, giving seven signed states including zero; the $T=6$ configuration extends this to magnitudes from 0 to 7 and fifteen signed states. The scale controls their spacing in the original activation domain. This provides the representation budgets used in the workload and Static-QAT experiments.

{BBS therefore combines uniformly spaced reconstruction values with input regions determined by the spike recurrence. A static uniform quantizer can use the same signed values while assigning inputs through uniformly spaced thresholds. Matching the number of states compares equal-sized output alphabets; the downstream experiment examines how these different assignments perform within the complete model.}

\subsection{Spiking Attention Mechanism}
\label{sec:spiking-attention}

\textbf{Spiking QKV representation.}
\label{sec:qkv-spiking}
BBS encodes the outputs of the Query, Key, and Value projections. For $T=4$, define the signed integer tensor $\widetilde Q$ by
\begin{equation}
  \mathrm{BBS}(Q)=s_q\widetilde Q,
  \qquad
  \widetilde Q=Q_{P1}+2Q_{P2}-Q_{N1}-2Q_{N2},
  \label{eq:bbs-q}
\end{equation}
where $Q_{Pj}$ and $Q_{Nj}$ are binary positive and negative tensors with magnitude weight $j$. Defining $\widetilde K$ and $\widetilde V$ analogously gives
\begin{equation}
  \mathrm{BBS}(Q)\,\mathrm{BBS}(K)^{T}
  =s_qs_k\,\widetilde Q\widetilde K^{T}.
  \label{eq:qkt-bbs}
\end{equation}
Expanding the binary planes expresses the product through binary interactions, signed accumulations, and power-of-two shifts, followed by the scale factor. In particular, multiplying two binary-weight contributions produces another power-of-two coefficient. Their magnitudes can therefore be handled by shifts, while the positive and negative pathways determine the accumulation sign. The same signed representation is retained for the Value projection.

\textbf{Spiking approximation of Softmax.}
\label{sec:spiking-softmax}
The identity $e^{x/\sqrt d}=2^{x/(\sqrt d\ln2)}$ converts exponential evaluation into an integer-exponent mapping. Using the threshold search above, UBS approximates the exponent by
\begin{equation}
  \begin{aligned}
    c(x)&=\mathrm{UBS}(x)=\sum_{t=1}^{U}2^{U-t}S(t)
    \approx\left\lfloor\frac{x}{\sqrt d\ln2}\right\rfloor,\\
    e^{x/\sqrt d}&\approx 2^{c(x)}
    =\prod_{t=1}^{U}2^{\,2^{U-t}S(t)}.
  \end{aligned}
  \label{eq:ubs-exp-approx}
\end{equation}
Each active bit contributes a power-of-two shift, and the integer exponent determines the combined shift amount. In this construction, UBS approximates the exponent code; the subsequent power-of-two mapping supplies the nonlinearity used for attention scores. Let
$C=\mathrm{UBS}(s_qs_k\widetilde Q\widetilde K^{U})$ and
$E=2^C$, with UBS and exponentiation applied elementwise.
The normalized attention output is
\begin{equation}
  \mathrm{Softmax}\!\left(\frac{QK^{U}}{\sqrt d}\right)V
  \approx \operatorname{diag}(E\mathbf{1})^{-1}E\,s_v\widetilde V,
  \label{eq:spiking-attention}
\end{equation}
where $\mathbf{1}$ sums over keys. Integer codes support shift-based exponentiation and accumulation with BBS-encoded values; row normalization and scale application remain floating-point operations. At inference, $s_qs_k$ can be absorbed into the UBS thresholds, leaving $s_v$ on the output.

{\textbf{Why evaluate attention-output fidelity?}
For a reference attention row $p_j\propto 2^{u_j}$, exponent-code errors $\delta_j$ yield
\begin{equation}
  \widehat p_j
  =\frac{p_j\,2^{\delta_j}}{\sum_k p_k\,2^{\delta_k}},
  \label{eq:attention-code-error}
\end{equation}
where the sum is over valid keys. A common exponent offset cancels during normalization, whereas differences between code errors redistribute attention across tokens. The effect on the output also depends on the associated Value vectors. This motivates measuring $\widehat P V$ against reference attention outputs in addition to fitting the exponent codes: the former evaluates the quantity passed to the next layer.}

For training, we use a straight-through estimator (STE)~\citep{STE} for spike firing. UBS and BBS return decoded integer representations without explicitly unrolling spike propagation through subsequent layers; the trained parameters are used for the corresponding temporal binary encoding at inference.

\subsection{Threshold Search and Differentiable Training}
\label{sec:training-encoding}

To approximate a target function $f(x)$, we optimize four thresholds $\theta=[\theta_1,\ldots,\theta_4]$ and $\alpha$ for the decoded output 
$P(x;\theta,\alpha)$:
\begin{equation}
  \min_{\theta,\,0<\alpha\leq1}\;
  \mathcal{L}(\theta,\alpha)
  =\frac{1}{N}\sum_{i=1}^{N}
  [P(x_i;\theta,\alpha)-f(x_i)]^2+\lambda\mathbb{I}.
  \label{eq:loss}
\end{equation}
Inputs $x_i$ are uniformly sampled from the target domain; $\mathbb{I}$ indicates violations of threshold ordering and range constraints, penalized by $\lambda$. Differential evolution~\citep{storn1997differential} provides a global approximate solution, followed by local grid refinement. The resulting parameters are used for the selected activation and exponential mappings. The global stage searches across threshold configurations, while local refinement adjusts the best candidate's partitions. Readout weights remain powers of two throughout the search, so fitting the target mapping does not change the arithmetic structure of the resulting code.

\textbf{Offline search.}
For a chosen target $f$ and domain $[a,b]$, offline search uses 1,201 uniformly spaced inputs, a population of 125 candidates, and 120 generations. Differential mutation samples $F\sim\mathcal U(0.5,1)$ and binomial crossover uses probability 0.7, with seed 42. Threshold bounds are $[1,60]$ and decay bounds are $[0.01,1]$. After differential evolution, thresholds are rounded to integers and locally refined. The target domain and threshold--weight association are specified for each mapping. This offline function fit supplies encoder parameters; the data-based recalibration in Section~\ref{sec:encoding-attention} is a separate mechanism experiment.

\begin{algorithm}[tbp]
\caption{Differential evolution and local refinement for UBS}
\label{alg:differential-evolution}
\begin{algorithmic}[1]
\Require Target $f$, sampled inputs $\{x_k\}$, feasible set $\mathcal B$, population size $N_P$, generations $G$
\Ensure Thresholds and decay $z^\star=(\theta^\star,\alpha^\star)$
\State Initialize $z_1,\ldots,z_{N_P}\in\mathcal B$; evaluate the loss in Equation~\ref{eq:loss}
\For{$g=1,\ldots,G$}
  \State Let $z_{\mathrm{best}}$ be the lowest-loss candidate
  \For{each candidate $z_i$}
    \State Select distinct $r_1,r_2\ne i$ and sample $F\sim\mathcal U(0.5,1)$
    \State Form $v_i=z_{\mathrm{best}}+F(z_{r_1}-z_{r_2})$
    \State Cross $v_i$ with $z_i$ at rate $0.7$, retaining at least one coordinate from $v_i$
    \State Repair the trial to satisfy $\mathcal B$ and evaluate its loss
    \State Retain the trial if its loss is no larger than that of $z_i$
  \EndFor
\EndFor
\State Round the best thresholds to integers and refine locally under the same constraints
\State \Return the best feasible $z^\star$
\end{algorithmic}
\end{algorithm}

\textbf{Surrogate gradients.}
Network training uses straight-through or smooth surrogate derivatives for discrete firing~\citep{STE}. For the attention UBS code, the clipped surrogate is
$Q_{\mathrm{att}}(x)=\operatorname{clip}(x/(\sqrt d\ln2),0,n)$;
for post-activation FFN UBS, it is
$Q_{\mathrm{ffn}}(x)=\operatorname{clip}(x,0,n)$, where $n=2^U-1$ is the maximum unsigned code. These surrogates propagate gradients through the corresponding code range while the forward computation retains discrete values.

For BBS, write $u=|x|/s$ and $y=s\,\operatorname{sign}(x)q(u)$, where $q$ is the decoded magnitude from the complete residual recurrence. If $q'_{\mathrm{sur}}(u)$ denotes its surrogate derivative, the scale gradient follows the chain rule:
\begin{equation}
 \frac{\partial\mathcal L}{\partial s}
 \approx
 \frac{\partial\mathcal L}{\partial y}\,
 \operatorname{sign}(x)
 \left[q(u)-u\,q'_{\mathrm{sur}}(u)\right].
 \label{eq:scale-gradient}
\end{equation}
The first term accounts for rescaling the decoded magnitude, and the second accounts for moving the normalized input relative to the thresholds. Smooth spike derivatives provide gradients near decision boundaries and are propagated through the residual updates. A smaller scale increases the normalized magnitude and threshold crossings; a larger scale widens the represented activation range. Learning $s$ therefore adapts the fixed magnitude codebook to each encoder's feature distribution.

\section{Results}
\label{sec:results}

\textbf{Benchmarks and metrics.}
We evaluate eight GLUE datasets spanning grammatical acceptability (CoLA), sentiment classification (SST-2), paraphrase identification (QQP and MRPC), semantic similarity (STS-B), and inference (MNLI, QNLI, and RTE). MNLI provides matched and mismatched development splits, giving nine reported task columns. Classification tasks use accuracy, CoLA uses Matthews correlation, and STS-B uses Spearman correlation, all on a 0--100 scale. The all-task average is the arithmetic mean of these nine columns. The reduced average excludes CoLA and follows the MNLI-split convention marked in Table~\ref{tab:glue-results}.

\textbf{Training setup.}
We evaluate Spora on GLUE~\citep{wang2018glue} after pre-training on English Wikipedia~\citep{devlin-etal-2019-bert} and BookCorpus~\citep{zhu2015aligning} at maximum sequence length 128. Training uses AdamW~\citep{loshchilov2018decoupled}, a learning rate of $5\times10^{-5}$, 5,000 warmup steps, and mixed precision~\citep{micikevicius2018mixed}. Multi-level BERT MLM distillation~\citep{sun-etal-2019-patient} transfers the BERT-base teacher's output distribution, hidden representations, and attention matrices, followed by GLUE fine-tuning. {Training uses four 24GB RTX4090 GPUs; layer-local mechanism evaluation uses one RTX4090 D.}

\subsection{Performance on GLUE}
\label{sec:baseline-comparison}

{
Table~\ref{tab:glue-results} compares Spora with full-precision BERT~\citep{devlin-etal-2019-bert}; quantized models Q2BERT~\citep{Q2BERT}, BinaryBERT~\citep{bai-etal-2021-binarybert}, BiBERT~\citep{BiBERT}, TernaryBERT~\citep{zhang-etal-2020-ternarybert}, BiT~\citep{BiT}, and BiPET~\citep{BiPFT}; and spiking language models. Among the quantized and spiking baselines, all except Q2BERT, WE-SpikingFormer, and SpikingFormer employ knowledge distillation.
}

\begin{table*}[tbp]
\centering
\caption{GLUE development results (0--100). QQP/MRPC: accuracy;
STS-B: Spearman's correlation; CoLA: MCC. Avg.\ excludes CoLA;
Avg.$_{\mathrm{all}}$ includes it. Asterisks indicate averages using
only one MNLI split. Time denotes time steps.
Spora$_{T=4}$'s QQP/MRPC F1 scores are 85.9/86.6; the reduced average for Spora$_{T=6}$ uses MNLI-m.}
\label{tab:glue-results}
\begin{adjustbox}{max width=\linewidth}
\begin{minipage}{173.8mm}
\centering

\setlength{\tabcolsep}{1.5pt}
\begin{tabular}{@{}l*{11}{c}@{}}
\toprule
Model & Time & QQP &
\shortstack{MNLI$_{m/mm}$} &
SST-2 & QNLI & RTE & MRPC & STS-B &
Avg. & CoLA & Avg.$_{\mathrm{all}}$ \\
\midrule

\rowcolor{gray!15}
\multicolumn{12}{l}{\textit{ANN}} \\
BERT
& -- & 91.3 & 83.8/83.4 & 92.3 & 90.7
& 69.3 & 88.2 & 89.4 & 86.1 & 60.0 & 83.2 \\
Q2BERT
& -- & 67.0 & 47.2/47.3 & 80.6 & 61.3
& 52.7 & 68.4 & 4.4 & 53.6 & 0.0 & 47.7 \\
BinaryBERT
& -- & 79.9 & 62.7/63.9 & 82.5 & 52.6
& 52.7 & 68.3 & 6.5 & 58.6 & 14.6 & 53.7 \\
BiBERT
& -- & 83.7 & 66.1/67.5 & 88.7 & 72.6
& 57.4 & 72.5 & 33.6 & 67.8 & 25.4 & 63.0 \\
TernaryBERT
& -- & 63.1 & 40.3/40.0 & 80.7 & 50.0
& 54.5 & 68.3 & 12.4 & 51.2 & 0.0 & 45.5 \\
BiT
& -- & 82.9 & 77.1/77.5 & 87.7 & 85.7
& 58.8 & 79.7 & 71.1 & 77.6 & 25.1 & 71.7 \\
BiPET
& -- & 83.7 & 69.5/70.6 & 86.2 & 81.7
& 66.1 & 76.2 & 80.2 & 76.8 & 22.9 & 70.8 \\

\midrule
\rowcolor{gray!15}
\multicolumn{12}{l}{\textit{SNN}} \\
SpikingFormer
& 4 & 83.8 & 67.8/68.6 & 82.7 & 74.6
& 58.8 & 74.0 & 72.3 & 73.4 & 16.7 & 66.6 \\
SpikeLM
& 4 & 87.9 & 76.0/76.9 & 86.5 & 84.9
& 65.3 & 78.7 & 84.3 & 80.1 & 37.9 & 75.4 \\
Sorbet
& 16 & 86.5 & 77.3/-- & 90.4 & 86.1
& 60.3 & 79.9 & 78.1 & 79.8$^{*}$ & -- & -- \\
SpikingBERT
& 125 & 86.8 & 78.1/-- & 88.2 & 85.2
& 66.1 & 79.2 & 82.2 & 80.8$^{*}$ & -- & -- \\
WE-SpikingFormer
& 4 & 85.1 & --/70.1 & 89.0 & 77.5
& 55.9 & 81.6 & 42.8 & 71.7 & 27.9 & 66.2 \\

\midrule
Spora$_{T=4}$
& 4 & 89.2 & 79.2/79.4 & 89.6 & 85.4
& 60.0 & 80.2 & 82.5 & 80.9$^{*}$ & 44.1 & 76.6 \\
Spora$_{T=6}$
& 6 & \textbf{89.4} & \textbf{80.1/80.6} & 89.9 & \textbf{86.9}
& 63.2 & \textbf{82.2} & \textbf{84.3} & \textbf{82.3}$^{*}$
& \textbf{47.4} & \textbf{78.2} \\
\bottomrule
\end{tabular}

\end{minipage}
\end{adjustbox}
\end{table*}

{
With four time steps, Spora reaches 76.6 average score across all reported GLUE columns and 80.9 under the table's reduced averaging convention. Relative to SpikeLM~\citep{xing2024spikelmgeneralspikedrivenlanguage}, it improves QQP by 1.3 points and MNLI-m/mm by 3.2/2.5 points. The largest gain is on CoLA: 44.1 MCC versus 37.9 for SpikeLM, 27.9 for WE-SpikingFormer~\citep{WE-SpikingFormer}, and 16.7 for SpikingFormer~\citep{zhou2025spikingformerkeyfoundationmodel}. On SST-2, Spora reaches 89.6\%, approaching Sorbet's~\citep{tang2025sorbet} 90.4\% with four rather than sixteen steps. It also exceeds SpikingBERT~\citep{SpikingBERT} on QQP, MNLI-m, and SST-2 with four rather than 125 steps. Spora also reaches 85.4\% on QNLI and 80.2\% on MRPC, compared with 84.9\% and 78.7\% for SpikeLM. On QQP, its 89.2\% accuracy is within 2.1 points of full-precision BERT. These improvements span sentence-pair classification, sentiment analysis, and grammatical acceptability, showing that the short-window design benefits several language-understanding tasks rather than a single benchmark.
}

\subsection{Encoding and Attention Analysis}
\label{sec:encoding-attention}

\textbf{Conditional-decay protocol.}
We evaluate Spora$_{T=6}$ on CoLA at maximum length 128,
comparing original four-step UBS, no-decay UBS ($\alpha=1$)
with recalibrated thresholds, jointly calibrated UBS,
and static uniform integer encoding.
Calibration minimizes exponent-code MSE on 256 training
examples under the same maximum search budget;
evaluation uses all 1,043 development examples in FP32,
excluding padding.

For attention scores $Z$, let $r=\operatorname{round}(Z)$.
The encoder input is $x=\operatorname{clip}(r,0,60)$,
and the calibration target is $x/(8\ln 2)$.
We compare $\widehat P V$ with two reference outputs:
\begin{equation}
  \left\{
  \begin{aligned}
    P_{\mathrm{raw}}V
      &=\operatorname{Softmax}(Z/8)V,\\
    P_{\mathrm{shared}}V
      &=\operatorname{Softmax}(x/8)V.
  \end{aligned}
  \right.
  \label{eq:conditional-decay-references}
\end{equation}
The raw-score reference measures total error, while the
shared-preprocessing reference measures coding error
after the same preprocessing.
For either reference $P$, output NMSE is
\begin{equation}
  \mathrm{NMSE}
  =\frac{\sum\|(\widehat P-P)V\|_F^2}
         {\sum\|PV\|_F^2},
  \label{eq:conditional-decay-nmse}
\end{equation}
where both sums are aggregated over valid queries.

\begin{figure*}[tbp]
  \centering
  \includegraphics[width=\textwidth]{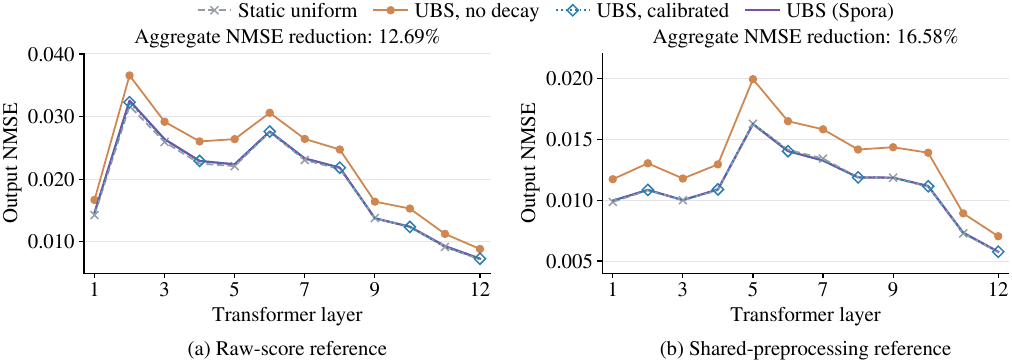}
  \caption{Attention-output NMSE for four-step integer
  encoders: (a) raw-score and (b) shared-preprocessing references.
  Annotations report aggregate reductions relative to recalibrated
  no-decay UBS.}
  \label{fig:encoding-attention}
\end{figure*}

\textbf{Attention-output fidelity.}
Original UBS lowers error relative to recalibrated
no-decay UBS in all 12 layers
(Figure~\ref{fig:encoding-attention}).
Aggregate NMSE decreases from 0.02100 to 0.01834
under the raw-score reference and from 0.01277 to
0.01066 under the shared-preprocessing reference,
corresponding to reductions of 12.69\% and 16.58\%,
respectively.
Static uniform encoding achieves raw-score NMSE
of 0.01805 and shared-preprocessing NMSE of 0.01063.
Joint UBS recalibration improves the original
by less than 0.2\% relative.

Conditional decay thus refines input partitions
while retaining binary readout, bringing temporal
encoding close to direct integer encoding in
attention fidelity.
The raw and shared references distinguish errors
introduced before encoding from those of the
integer mapping itself.
Improvement under both references connects the
residual update to the attention output, while
the small recalibration gain indicates that the
checkpoint's UBS configuration is already close
to the jointly calibrated setting.

{
\textbf{Comparison with static activation quantization.}
\label{sec:static-qat-comparison}
We next compare the complete BBS/UBS design against Static-QAT on CoLA.

\begin{table}[tbp]
\centering
\caption{{CoLA comparison under matched
representation budgets: 15 signed states and four-bit UBS.}}
\label{tab:static-qat-comparison}

{
\setlength{\tabcolsep}{3pt}
\begin{tabular*}{\linewidth}{
  @{\extracolsep{\fill}}lrr@{}
}
\toprule
Method &
\shortstack{Acc.\\(\%)} &
\shortstack{MCC\\$\times100$} \\
\midrule
Static-QAT    & 76.61          & 40.97 \\
Spora$_{T=6}$ & \textbf{79.00} & \textbf{47.40} \\
\bottomrule
\end{tabular*}
}
\end{table}

Spora$_{T=6}$ uses three magnitude steps $[4,2,1]$ with 15 signed states; Static-QAT uses symmetric 4-bit quantization with the same 15 states. Both have a four-bit UBS representation budget. Static-QAT replaces the BBS and UBS encoders with uniform quantizers with learnable scales, retaining the architecture and base-2 attention normalization. It follows the four-epoch MLM distillation schedule with the same BERT-base teacher, temperature 2.0, and MLM/hidden/attention loss weights of 1.0/0.5/5.0, followed by CoLA fine-tuning and development-MCC checkpoint selection. Its best checkpoint is evaluated alongside the existing fine-tuned Spora$_{T=6}$ checkpoint.

Table~\ref{tab:static-qat-comparison} reports 47.40 MCC and 79.00\% accuracy for Spora$_{T=6}$, exceeding Static-QAT by 6.43 MCC points and 2.39 accuracy points. The two comparisons address successive parts of the design. The layer-local study measures how UBS partitions affect a specific attention operator and shows fidelity close to static integer encoding. The downstream comparison includes signed BBS features, UBS mappings, and their interaction with training. Its 6.43-point MCC gain establishes a task-level advantage for the complete design under the matched representation budget.
}

\subsection{Computational Cost and Numerical Precision}
\label{sec:energy-analysis}
\label{sec:hardware-oriented-evaluation}

\textbf{Arithmetic energy accounting.}
We estimate energy as operation counts multiplied by technology-level unit costs~\citep{Horowitz}. For FP32 arithmetic, the reference costs are 4.6~pJ per MAC, 3.7~pJ per multiplication, and 0.9~pJ per addition. In the FP32 linear-layer accounting, we use the addition cost as a proxy for a shift. The low-bit attention model uses 0.03~pJ for an 8-bit integer addition and 0.1~pJ for a 32-bit integer addition; exponential evaluation uses the 1.7~pJ estimate adopted in Sorbet~\citep{tang2025sorbet}. These are arithmetic energy estimates for the specified operation models.

\begin{figure*}[tbp]
  \centering
  \includegraphics[width=\linewidth]{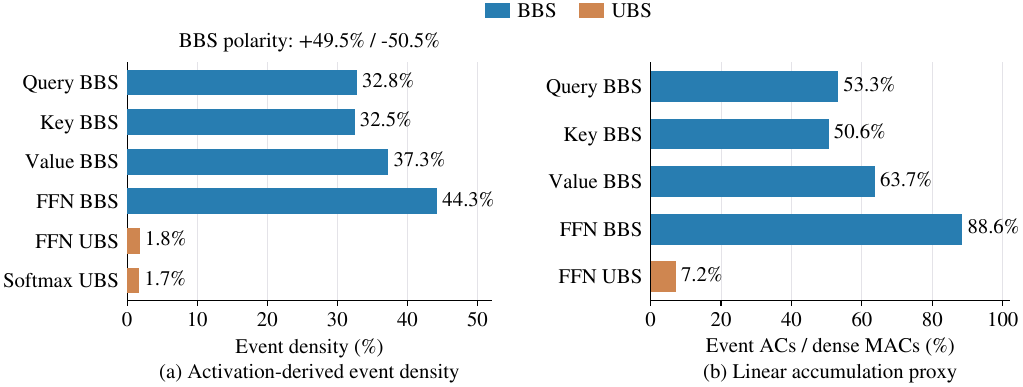}
  \caption{{Activation-derived workload on CoLA at padded length 128. Left: event density by component. Right: estimated event-driven accumulations relative to dense MACs in spike-to-linear paths.}}
  \label{fig:component-workload}
\end{figure*}

For a linear map with dense MAC count $N_{\mathrm{dense}}$, let $r_t$ be the active fraction in signed bit plane $t$ and $w_t$ its numerical weight. Event-wise accounting gives
\begin{equation}
\begin{aligned}
 N_{\mathrm{AC}}&=N_{\mathrm{dense}}\sum_t r_t,\\
 N_{\mathrm{shift}}&=N_{\mathrm{dense}}
                       \sum_{t:\,|w_t|>1}r_t,\\
 E_{\mathrm{event}}&=N_{\mathrm{AC}}E_{\mathrm{AC}}
                     +N_{\mathrm{shift}}E_{\mathrm{shift}},\\
 E_{\mathrm{dense}}&=N_{\mathrm{dense}}E_{\mathrm{MAC}}.
\end{aligned}
\label{eq:event-energy}
\end{equation}
The shifted planes are identified by weight magnitude rather than their storage order. For weights $[2,1]$, only magnitude-two events contribute to the shift term. Sign selects addition or subtraction. This notation connects the energy estimate to the measured event counts in Figure~\ref{fig:component-workload}.

Attention is accounted for component by component: binary-plane products and shifts for $QK^T$, UBS code generation and shifts for exponentiation, row normalization, accumulation with $V$, and output scaling. The energy of a component is $\sum_o N_o E_o$ over its arithmetic operations. Normalization and value scaling remain explicit in Table~\ref{tab:energy-fixedpoint}; the total reduction therefore refers to the complete core-attention accounting shown there.

\textbf{Theoretical attention energy.}
For BERT-base at sequence length 128, with 12 heads of dimension 64, estimated core attention energy decreases from 116.272 to 1.7083~$\mu$J per layer, a 98.53\% reduction (Table~\ref{tab:energy-fixedpoint}, left). The savings arise from accumulation-and-shift computation for $QK^T$ and $AV$ and integer-exponent mappings for Softmax. Query/key scales are absorbed into UBS thresholds; value scaling is applied to the output.

\begin{table*}[tbp]
\centering

\caption{Left: theoretical attention energy in $\mu$J
(Conv.: conventional; Approx.: proposed).
{Right: fixed-point emulation on CoLA.
State and Prob. denote BBS membrane-state and
attention-probability formats; Sat. is the percentage
of clipped state values.}}
\label{tab:energy-fixedpoint}
\begin{adjustbox}{max width=\linewidth}
\begin{minipage}{173.8mm}
\centering

\begin{minipage}[c]{0.37\linewidth}
\centering
\setlength{\tabcolsep}{2pt}
\begin{tabular*}{\linewidth}{
  @{\extracolsep{\fill}}lrr@{}
}
\toprule
Module & Conv. & Approx. \\
\midrule
$QK^{T}$                & 57.881  & 0.303 \\
$e^{x}/2^{\mathrm{UBS}}$ & 0.334   & 0.028 \\
Denominator             & 0.176   & 0.005 \\
$AV$                    & 57.881  & 1.008 \\
$s_v$ scaling           & --      & 0.364 \\
\midrule
Total                   & 116.272 & 1.7083 \\
\bottomrule
\end{tabular*}
\end{minipage}%
\hspace{0.02\linewidth}%
\begin{minipage}[c]{0.61\linewidth}
\centering

\setlength{\tabcolsep}{2pt}
\begin{tabular*}{\linewidth}{
  @{\extracolsep{\fill}}lccrrr@{}
}
\toprule
Format & State & Prob. &
\shortstack{Acc.\\(\%)} &
\shortstack{MCC\\$\times100$} &
\shortstack{Sat.\\(\%)} \\
\midrule
FP32   & Float & Float & 77.95 & 44.39 & -- \\
16-bit & Q6.10 & U16   & 77.56 & 43.44 & 0.0194 \\
12-bit & Q6.6  & U12   & 76.51 & 40.64 & 0.0193 \\
8-bit  & Q6.2  & U8    & 77.28 & 42.90 & 0.0184 \\
\bottomrule
\end{tabular*}
\end{minipage}
\end{minipage}
\end{adjustbox}
\end{table*}

{
\textbf{Workload and precision protocol.}
We use the 1,043-example CoLA development set at maximum length 128 and the fine-tuned four-step model: two BBS magnitude steps $[2,1]$, seven signed states $\{-3,-2,-1,0,1,2,3\}$, and four UBS bit-planes.

\textbf{Event workload.}
Figure~\ref{fig:component-workload} aggregates active events and temporal slots across layers. BBS densities range from 32.5\% to 44.3\%, with balanced positive/negative events (49.5\%/50.5\%); FFN and attention UBS densities are 1.8\% and 1.7\%. For spike-to-linear paths, active-event counts multiplied by output fanout estimate event-driven accumulations. Their aggregate ratio to dense MACs is 50.1\%, falling to 7.2\% for FFN UBS, which offers the sparsest accumulation path. The contrast with FFN BBS's 88.6\% ratio shows where event sparsity contributes most. Binary weights determine how an event is processed, while activation-dependent firing determines how many such operations are needed; these are complementary sources of computational savings.
}

{
\textbf{Fixed-point emulation.}
We quantize BBS membrane potentials to $Q6.(b-6)$, round the decay coefficient, and uniformly quantize attention probabilities to $b\in\{8,12,16\}$ bits. Integer thresholds remain exact; learned scales and other operators remain floating point, and automatic mixed precision is disabled. Table~\ref{tab:energy-fixedpoint} (right) reports 43.44/42.90 MCC at 16/8 bits, decreases of 0.95/1.48 points from FP32's 44.39. The 12-bit configuration reaches 40.64 MCC; saturation remains below 0.02\% in all cases. The 16- and 8-bit formats therefore retain most downstream accuracy for low-precision states and probabilities.
}

\textbf{Whole-model comparison.}
The FP32 MAC/AC accounting summarized in Table~\ref{tab:model-energy} estimates 11.17~mJ for four-step Spora, compared with 51.41~mJ for BERT-base, a ratio of approximately $4.60\times$. This whole-model estimate includes computation outside the core attention block and therefore has a different scope from the 98.53\% core-attention reduction.

\begin{table*}[t]
\centering
\caption{Whole-model arithmetic energy estimates at the reported configurations.
Starred values are inferred from energy ratios in the corresponding papers.
The accounting uses FP32 MAC/AC reference costs.}
\label{tab:model-energy}
\begin{adjustbox}{max width=\linewidth}
\begin{minipage}{173.8mm}
\centering

\setlength{\tabcolsep}{2pt}
\begin{tabular*}{\linewidth}{@{\extracolsep{\fill}}lccccccc@{}}
\toprule
Model
& BERT-base
& LIF-BERT$_{T=4}$
& SpikeLM$_{T=4}$
& SpikingBERT
& SpikeBERT$_{T=4}$
& Sorbet$_{T=16}$
& Spora$_{T=4}$ \\
\midrule
Energy (mJ)
& 51.41
& 7.98
& 13.74
& 12.85$^{*}$
& 14.74
& 24.14$^{*}$
& \textbf{11.17} \\
\bottomrule
\end{tabular*}
\end{minipage}
\end{adjustbox}
\end{table*}

\begin{table*}[t]
\centering

\caption{Ablation study on GLUE.
Spora$_1$ removes BBS encoding from the Query-layer input.
Spora$_{\mathrm{Softmax}}$ restores standard scaled
dot-product attention while retaining BBS-encoded Q/K/V.
Spora w/o KD removes knowledge distillation.}
\label{tab:ablation}
\begin{adjustbox}{max width=\linewidth}
\begin{minipage}{173.8mm}
\centering

\setlength{\tabcolsep}{2.5pt}
\begin{tabular}{@{}l*{9}{c}@{}}
\toprule
Model & QQP &
\shortstack{MNLI$_{\mathrm{m/mm}}$} &
SST-2 & QNLI & RTE & MRPC & STS-B & CoLA & Avg. \\
\midrule
Spora
& 89.2 & 79.2/79.4 & 89.6 & 85.4
& 60.0 & 80.2 & 82.5 & 44.1 & 76.6 \\
Spora$_1$
& 88.9 & 79.3/79.4 & 89.4 & 85.2
& 58.6 & 80.9 & 83.2 & 46.8 & 76.8 \\
Spora$_{\mathrm{Softmax}}$
& 88.8 & 78.5/79.3 & 89.7 & 85.0
& 62.1 & 81.2 & 80.7 & 45.2 & 76.7 \\
Spora w/o KD
& 87.1 & 76.4/76.7 & 87.2 & 83.2
& 57.1 & 79.4 & 79.7 & 38.5 & 73.9 \\
\bottomrule
\end{tabular}
\end{minipage}
\end{adjustbox}
\end{table*}

\begin{figure*}[t]
  \centering
  \includegraphics[width=1\linewidth]{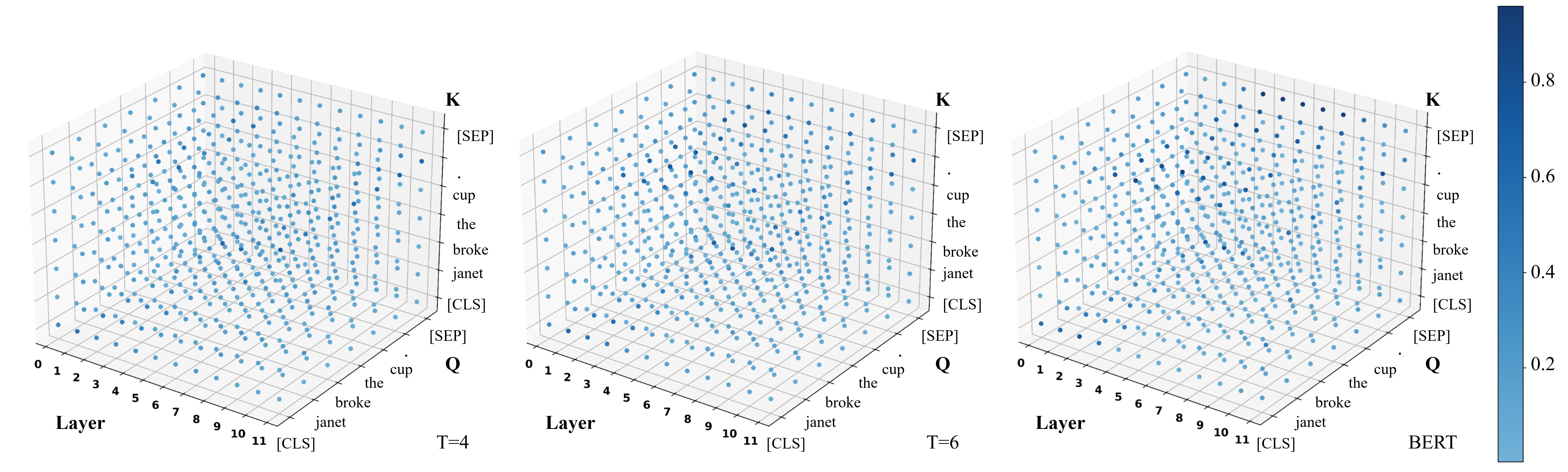}
  \caption{Attention score visualization of BERT and Spora with time steps 4 and 6 on the CoLA sample ``Janet broke the cup.'' across 12 layers. $Q$ denotes the query token and $K$ denotes the key token.}
  \label{fig:attention-visualization}
\end{figure*}

\subsection{Ablation Study}
\label{sec:ablation-study}

All variants in Table~\ref{tab:ablation} are pre-trained on the same corpus and fine-tuned on GLUE.
{Replacing approximate attention with standard Softmax changes the average score from 76.6 to 76.7; removing Query-input BBS changes it to 76.8. The complete model thus retains comparable average accuracy with both encoding and approximate attention enabled. These ablations answer complementary questions: Query-input BBS tests the effect of encoding features before the projection, whereas the Softmax variant tests the nonlinear attention approximation with BBS-encoded Q/K/V retained. The small average differences show that both choices preserve task performance within this architecture. Under the same time-step and training budgets, knowledge distillation increases the average from 73.9 to 76.6 and CoLA MCC from 38.5 to 44.1.}




\textbf{Increasing the temporal budget.}
As shown in Table~\ref{tab:glue-results}, Spora$_{T=6}$ improves the all-task average from 76.6 to 78.2, a gain of 1.6 points, and improves every reported task column. CoLA rises by 3.3 points, RTE by 3.2, MRPC by 2.0, and STS-B by 1.8. The additional magnitude bit expands the signed codebook from seven to fifteen states while keeping four-stage UBS attention. These results expose a controllable accuracy--representation tradeoff: four-step Spora provides the shorter configuration, while six-step Spora uses finer signed representations to improve task performance.

\textbf{Attention distribution.}
Figure~\ref{fig:attention-visualization} compares BERT and Spora across 12 layers. Spora exhibits more dispersed attention than BERT on this example, with further dispersion at $T=6$. Extending magnitude weights from $[1,2]$ to $[1,2,4]$ increases representational resolution; the associated task improvements are quantified in Table~\ref{tab:glue-results}.


\section{Discussion and Conclusion}

Spora gives temporal spike sequences a numerical structure that can be used directly by attention operators. Binary weights preserve place values; conditional decay shapes input partitions; sign-separated BBS and learned scales accommodate the feature distributions of language models. The resulting design links representation to accumulation, shifts, and integer-exponent mappings rather than treating the spike encoder as an isolated discretization step.

The experiments connect these mechanisms at several levels. Conditional-decay analysis measures local attention fidelity, the matched-budget Static-QAT comparison evaluates the complete encoding design after training, and GLUE ablations characterize the accuracy cost of the selected projection and attention approximations. Extending BBS from four to six steps raises the all-task average from 76.6 to 78.2 and CoLA MCC from 44.1 to 47.4. Event workloads, arithmetic energy estimates, and fixed-point emulation further describe the computational properties of the representation.

An immediate extension is to combine activation encoding with weight quantization so that parameter storage and event arithmetic can be optimized together. Implementations that exploit signed bit planes and event sparsity would also allow the operation-level savings to be studied alongside memory traffic and execution overhead. The present results support a joint design principle: short-window codes should preserve useful semantic distinctions and expose numerical structure that makes the next operation simpler.




\bibliographystyle{cas-model2-names}
\bibliography{reference}
\end{document}